\documentclass[runningheads,a4paper]{llncs}

\usepackage{amssymb}
\usepackage{graphicx}

\usepackage{epstopdf}
\usepackage{subfigure}
\usepackage{amsmath}
\usepackage{algorithm}
\usepackage{algorithmic}
\usepackage[algo2e]{algorithm2e}

\usepackage{url}
\urldef{\mailsa}\path|Bo.Han@student.uts.edu.au.com|
\urldef{\mailsb}\path|{Ivor.Tsang,Ling.Chen}@uts.edu.au.com|

\begin{document}

\mainmatter  

\title{On the Convergence of A Family of Robust Losses for Stochastic Gradient Descent}

\titlerunning{On the Convergence of A Family of Robust Losses for Stochastic Gradient Descent}

%
%
\author{Bo Han%
\and Ivor W. Tsang\thanks{Ivor W. Tsang is the corresponding author.}\and Ling Chen}
\authorrunning{Bo Han et al.}

\institute{Centre for Quantum Computation and Intelligent Systems\\
University of Technology Sydney\\
\mailsa\\
\mailsb\\
}

%
%
\maketitle

\begin{abstract}
The convergence of Stochastic Gradient Descent (SGD) using convex loss functions has been widely studied. However, vanilla SGD methods using convex losses cannot perform well with noisy labels, which adversely affect the update of the primal variable in SGD methods.
Unfortunately, noisy labels are ubiquitous in real world applications such as crowdsourcing. To handle noisy labels, in this paper, we present a family of robust losses for SGD methods.
By employing our robust losses, SGD methods successfully reduce negative effects caused by noisy labels on each update of the primal variable. We not only reveal that the convergence rate is $\mathcal{O}(1/T)$ for SGD methods using robust losses, but also provide the robustness analysis on two representative robust losses. Comprehensive experimental results on six real-world datasets show that SGD methods using robust losses are obviously more robust than other baseline methods in most situations with fast convergence.
\end{abstract}

\section{Introduction}
To handle large-scale optimization problems, a popular strategy is to employ Stochastic Gradient Descent (SGD) methods because of two advantages. First, they do not need to compute all gradients over the whole dataset in each iteration, which lowers computational cost per iteration. Secondly, they only process a mini-batch of data
points \cite{cotter2011better} or even one data point \cite{mitliagkas2013memory} in each iteration,
which vastly reduces the memory storage. Therefore, many researchers have extensively
studied and applied various SGD methods \cite{nesterov2012efficiency,agarwal2013stochastic,cjh15akk}. For instance, Large-Scale SGD \cite{bottou-2010} has been substantially applied to the
optimization of deep learning models \cite{LeNCLPN11}. Primal Estimated Sub-Gradient Solver (Pegasos)
\cite{ShalevShwartzSSC11} is employed to speed up the Support Vector Machines (SVM) methods, which is suitable for large-scale text classification problems.

\begin{figure}[!tp]
\vskip -0.1in
\begin{center}
\centerline{\includegraphics[width=0.9\textwidth]{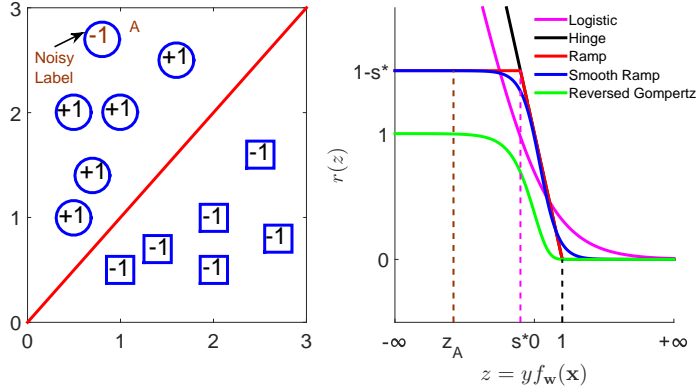}}
\caption{\textbf{Left Panel:} Squares represent real negative instances. Circles denote real positive instances, however one circle instance ``A'' is erroneously annotated as negative class, which creates a noisy label. \textbf{Right Panel:} Red curve and blue curve respectively denote Ramp Loss and Smooth Ramp Loss parameterized by $s^*$. Magenta curve, black curve and green curve correspond to Logistic Loss, Hinge Loss and Reversed Gompertz Loss accordingly. It can be observed that the incorrectly labeled instance ``A'' in the left panel can be regarded as the outlier of negative class, and its loss value $r(z_{A})$ is upper bounded by Ramp Loss, Smooth Ramp Loss, and Reversed Gompertz Loss (see ``$z_{A}$'' in the right panel).}
\label{Robust-Losses}
\end{center}
\vskip -0.5in
\end{figure}

However, vanilla SGD methods suffer from the label noise problem since the noisy labels adversely affect the update of the primal variable in SGD methods. Unfortunately, the label noise problems are very common in real-world applications. For instance, Amazon Mechanical Turk (MTurk) is a crowdsourcing Internet platform that takes advantage of human intelligence to provide supervision, such as labeling different kinds of bird pictures and annotating keywords according to geoscience records. However, the quality of annotations is not always satisfactory because many workers are not sufficiently trained to label or annotate such specific data \cite{yan2010modeling,BiWKT14}. Another situation is where the data labels are automatically inferred from user online behaviors or implicit feedback. For example, the existing recommendation algorithms usually consider a user clicking on an online item (e.g., advertisements on Youtube or eBay) as a positive label indicating user preference, whereas users may click the item for different reasons, such as curiosity or clicking by mistake. Therefore, the labels inferred from online behaviors are often noisy.

The aforementioned issues lead to a challenging question~- if the majority of data labels are incorrectly annotated, can we reduce the negative effects on SGD methods caused by these noisy labels? Our high-level idea is to design a robust loss function with a threshold for SGD methods. We illustrate our idea by using a binary classification example. In the left panel of \mbox{Figure~\ref{Robust-Losses}}, we notice that the instance $\mathbf{x}_{A}$ (i.e., data point ``A'') is incorrectly annotated with the label $y_{A} = -1$, which is opposite to its predicted label value (+1) according to the hyperplane. Moreover, this instance is far away from the distribution of negative class. Therefore, this instance $\mathbf{x}_{A}$ with the noisy label $y_{A}$ can be regarded as the outlier of negative class.

Let the output of the classifier $f_\mathbf{w}$ for a given $\mathbf{x}$ be $f_{\mathbf{w}}(\mathbf{x})$. Let $z$ be the product of the real label and the predicted label of an instance $\mathbf{x}$ (i.e., $z = yf_{\mathbf{w}}(\mathbf{x})$). Then, given the outlier $\{\mathbf{x}_A, y_A\}$ in the left panel of \mbox{Figure~\ref{Robust-Losses}}, we have $z_A = y_{A}f_{\mathbf{w}}(\mathbf{x}_{A}) < 0$. As illustrated in the right panel of Figure \ref{Robust-Losses}, with $z$ on the x-axis, the gradient of Hinge Loss is non-zero on the $z_A$, which will mislead the update of the primal variable $\mathbf{w}$ in SGD methods. However, if the loss function has a \mbox{threshold}, for example Ramp Loss \cite{collobert2006trading} in Figure~\ref{Robust-Losses} with a threshold $1-s^*$, the gradient of Ramp Loss on the $z_A$ is zero, which minimizes the negative effects caused by this outlier on the update. Therefore, it is reasonable to employ the loss with a threshold for SGD methods in the label noise problem.

Although the Ramp Loss is robust to outliers, it is computationally hard to optimize due to its nonsmoothness and nonconvexity \cite{YuYXWS10}. Therefore, we consider to relax the Ramp Loss into smooth and locally strongly-convex loss. With random initialization, SGD methods can converge into a qualified local minima with a fast speed. Our main contributions are summarized as follows.
\begin{enumerate}
\setlength{\parsep}{0pt}
\setlength{\itemsep}{0pt}
\item\label{con1} We present a family of robust losses, which specifically benefit SGD methods to reduce the negative effects introduced by noisy labels, even under a high percentage of noisy labels.
\item\label{con2} We reveal that the convergence rate is $\mathcal{O}(1/T)$ for SGD methods using the proposed robust losses. Moreover, we provide the robustness analysis on two representative robust losses.
\item\label{con3} Comprehensive experimental results on varying scale datasets with noisy labels show that SGD methods using robust losses are obviously more robust than other baseline methods in most situations with fast convergence.
\end{enumerate}

\section{Related Works}
First, our work is closely related to SGD methods. For example, Xu proposes the Averaged Stochastic Gradient Descent (ASGD) method \cite{xu2011towards} to lower the testing error rate of the SGD \cite{bottou-2010}. However, their work is based on the assumption that the data is clean, which significantly limits their applicability to the label noise problem. Ghahdimi \& Lan introduce a randomized stochastic algorithm to solve nonconvex problems \cite{ghadimi2013stochastic}, and then generalize the accelerated gradient method to improve the convergence rate if the problem is nonconvex \cite{ghadimi2015accelerated}. However they do not focus on learning with noisy labels specifically, and do not consider strongly convex regularizer.

Second, our work is also related to bounded nonconvex losses for robust classification. For example, Collobert et al. propose the bounded Ramp Loss for support vector machine (SVM) classification problems. Wang et al. further propose a robust SVM based on a smooth version of Ramp Loss for suppressing the outliers \cite{wang2008training}. Their models are commonly inferred by Concave-Convex Procedure (CCCP) \cite{collobert2006trading}. However, both of them do not consider that SGD methods suffer from the label noise problem. In other words, their works do not improve SGD methods using the smooth version of Ramp Loss in the label noise problem.

Finally, our work is highly related to noisy labels. For instance, Reed \& Sukhbaatar focus on training deep neural networks using noisy labels \cite{reed2014training,sukhbaatar2015training}. \mbox{Natarajan} et al. propose a probabilistic model for handling label noise problems \cite{natarajan2013learning}. However, all these works are unrelated to SGD methods. Moreover, they cannot be used in real-time or large-scale applications due to their high computational cost. It is also demonstrated that the 0-1 loss function is robust for outliers. However, the 0-1 loss is neither convex nor differentiable, and it is intractable for real learning
algorithms in practice. Even though the surrogates of 0-1 loss is convex \cite{bartlett2006convexity}, they are very sensitive to outliers. To the best of our knowledge, the problem of SGD methods for noisy labels has not yet been successfully addressed. This paper therefore studies this problem and provides an answer with theoretical analysis and empirical verification.

\section{A Family of Robust Losses for Stochastic Gradient Descent}
In this section, we begin with the definition of a family of robust losses for SGD methods. Under this definition, we introduce two representative robust losses: Smooth Ramp Loss and Reversed Gompertz Loss. Then, we reveal the convergence rate of SGD methods using robust losses, and provide the robustness analysis on two representative robust losses.
\subsection{Notations and Definitions}
Let $\mathcal{D} = \{\mathbf{x}_i, y_i\}_{i=1}^n $ be the training data, where $\mathbf{x}_i \in \mathbb{R}^d$ denotes the $i$th instance and $y_i \in \{-1,+1\}$ denotes its binary label. The basic support vector machine model for classification is represented as
\vspace{-0.8em}
\begin{equation}\label{obj-classification}
\min_{\mathbf{w}} G(\mathbf{w})= \min_{\mathbf{w}}\frac{1}{n}\sum_{i=1}^{n} g_i(\mathbf{w})
\end{equation}
where $\mathbf{w} \in \mathbb{R}^d $ is the primal variable. Specifically, $g_i(\mathbf{w}) = \rho_\lambda(\mathbf{w}) + r(\mathbf{w};\{\mathbf{x}_i,y_i\})$ where $\lambda$ is the regularization parameter, $\rho_\lambda(\mathbf{w})$ is the regularizer and $r(\mathbf{w};\{\mathbf{x}_i,y_i\})$ is a loss function.

Based on Restricted Strong Convexity (RSC) and Restricted Smoothness (RSM) \cite{agarwal2012fast,loh2015regularized}, we propose two extended definitions. We use $\lVert \cdot \rVert$ to denote the Euclidean norm, and $B_d(\mathbf{w^*},\gamma)$ to denote the $d$ dimensional Euclidean ball of radius $\gamma$ centered at local minima $\mathbf{w^*}$. And we assume that function $G$ and $g_i$ are continuously differentiable.

\begin{definition}\label{ARSC}
\textbf{(Augmented Restricted Strong Convexity (ARSC))} If there exists a constant $\alpha>0$ such that for any $\mathbf{w},\tilde{\mathbf{w}} \in B_d(\mathbf{w^*},\gamma)$, we have
\begin{equation}
G(\mathbf{w}) - G(\tilde{\mathbf{w}}) - \langle\nabla G(\tilde{\mathbf{w}}), \mathbf{w}-\tilde{\mathbf{w}}\rangle \geq \frac{\alpha}{2}\lVert \mathbf{w} - \tilde{\mathbf{w}} \rVert^2
\end{equation}
\end{definition}
\emph{then $G$ satisfies Augmented Restricted Strong Convexity.}\\
\begin{definition}\label{ARSM}
\textbf{(Augmented Restricted Smoothness (ARSM))} If there exists a constant $\beta>0$ such that for any $i \in \{1,\cdots,n\}$ and $\mathbf{w},\tilde{\mathbf{w}} \in B_d(\mathbf{w^*},\gamma)$, we have
\begin{equation}
g_i(\mathbf{w}) - g_i(\tilde{\mathbf{w}}) - \langle\nabla g_i(\tilde{\mathbf{\mathbf{w}}}), \mathbf{w}-\tilde{\mathbf{w}}\rangle \leq \frac{\beta}{2}\lVert \mathbf{w} - \tilde{\mathbf{w}} \rVert^2
\end{equation}
\end{definition}
\emph{then $g_i$ satisfies Augmented Restricted Smoothness.}\\
\vspace{-1.2em}

\subsection{A Family of Robust Losses}\label{Losses and Algorithm}
We first present the motivation and definition of a family of robust losses. Take Support Vector Machines (SVM) with convex hinge loss as an example. SGD methods are commonly used to optimize the SVM model for large-scale learning. However, if data points with noisy labels deviate significantly from the hyperplane greatly, these mislabeled data points can be equally viewed as outliers. These outliers will severely mislead the update of the primal variable in SGD methods. Therefore, it is intuitive to design a loss function with a threshold, which truncates the value that exceeds the threshold. Inspired by Ramp Loss \cite{collobert2006trading}, we consider whether we can design a family of bounded, locally strongly-convex and smooth losses. If we combine this new loss with strongly-convex regularizer, the objective then satisfies the ARSC (i.e., Def.~\ref{ARSC}) and ARSM (i.e., Def.~\ref{ARSM}) simultaneously. Here, we define a family of robust losses $r(z)$ for SGD methods, where $z$ is the variable of loss function in the x-axis of Figure \ref{Robust-Losses}.
\begin{definition}\label{def3} A loss function $r(z)$ is robust for SGD methods if it simultaneously meets the following conditions:
\begin{enumerate}
\setlength{\parsep}{0pt}
\setlength{\itemsep}{0pt}
\item\label{con1} Upper bound condition - it should be bounded such that $\lim\limits_{z\to-\infty}r'(z)= 0$.
\item\label{con2} Locally $\lambda$-strongly convex condition - it should be locally $\lambda$-strongly convex if there exists a constant $\lambda > 0$ such that
$r(z) - \frac{\lambda}{2}\lVert z \rVert^2$ is convex when $z \in B_1(z^*,\gamma)$, where $B_1(z^*,\gamma)$ denotes the 1 dimensional Euclidean ball of radius $\gamma > 0$ centered at local minima $z^*$.
\item\label{con3} Smoothly decreasing condition - it should be monotonically decreasing and continuously differentiable.
\end{enumerate}
\end{definition}
\begin{remark}
We explain three conditions in Definition \ref{def3}. \ref{con1}) Since the upper bound can be equally viewed as the threshold, it is natural that the negative effects introduced by outliers are removed by the upper bound. \ref{con2}) The loss function should be locally $\lambda$-strongly convex. If the loss function is locally $\lambda$-strongly convex and the regularizer is globally $\lambda$-strongly convex (e.g., $\frac{\lambda}{2}\lVert \mathbf{w} \rVert^2$), the objective $G(\mathbf{w})$ is locally strongly-convex. Then, objective $G(\mathbf{w})$ satisfies the ARSC. \ref{con3}) If the loss function is monotonically decreasing, we reasonably assume that the objective is non-increasing around some local minima, which is convenient to prove the convergence rate. If the loss function is differentiable at every point, $g_i(\mathbf{w})$ satisfies the ARSM when $\frac{\lambda}{2}\lVert \mathbf{w} \rVert^2$ is used.
\end{remark}
Then a family of robust losses for SGD methods can be acquired under these conditions. Here, we propose two representative robust losses that perfectly satisfy the above three conditions. Both of them are presented in Figure \ref{Robust-Losses} and employed through the whole paper.

The first one is the Smooth Ramp Loss \eqref{SRL}, which is the smooth version of Ramp Loss\footnote{The common optimization method for Ramp Loss is using Concave-Convex Procedure (CCCP). However, CCCP is time-consuming compared to SGD methods.}. If we smooth the Ramp Loss around $s^*$ and around 1, it is much easier to optimize and satisfy the ARSM. Therefore, we employ reversed sigmoid function to represent the Smooth Ramp Loss.
\begin{equation}\label{SRL}
r(s^*,z) = \frac{1-s^*}{1+e^{\alpha_{s^*}(z + \beta_{s^*})}}
\end{equation}
where we set the $s^*$ of Ramp Loss, then the parameters $\alpha_{s^*}$ and $\beta_{s^*}$ of Smooth Ramp Loss are determined by minimizing the difference between Smooth Ramp Loss and Ramp Loss.

The second one is the Reversed Gompertz Loss, which is a special case of the Gompertz function \cite{garg1970maximum} and we reverse the Gompertz function by the y-axis.
\begin{equation}
r(c^*,z) = e^{-e^{c^{*}\cdot z}}
\end{equation}
where the curve of this loss is controlled by parameter $c^{*}$. The aforementioned losses are integrated into the SVM model and SGD methods are employed to update the primal variable $\mathbf{w}$.

By employing two above robust losses, we finally summarize the robust SGD algorithm - Stochastic Gradient Descent with Robust Losses in Algorithm \ref{SGDRL}. Specifically, the generalized algorithm consists of two special cases. For Stochastic Gradient Descent with Smooth Ramp Loss, the algorithm employs \lq\lq Set~I and Update I\rq\rq. For Stochastic Gradient Descent with Reversed Gompertz Loss, the algorithm employs \lq\lq Set II and Update II\rq\rq. In practical implementations, we often choose option A and also provide averaging option B.

\begin{algorithm*}
\KwIn{$\lambda \geq 0$, $s^*, c^*$, the learning rate $\eta$, the max number of epochs $T_{max}$, and the training set $\mathcal{D} = \{\mathbf{x}_i, y_i\}_{i=1}^n $}
{\bfseries Initialize:} $\tilde{\mathbf{w}}^{(0)} = \mathbf{0}$

{\bfseries Set:} $\left\{
\begin{aligned}
I&: f(\alpha_{s^*},\beta_{s^*},g) = e^{\alpha_{s^*}(g+\beta_{s^*})}\\
II&: f(c^*,g) = c^* g - e^{c^* g}
\end{aligned}
\right.
$

\For{$epoch = 1,2,\dotsc,T_{max}$}{
{\bfseries Preprocess:} $\mathbf{w}^{(0)} = \tilde{\mathbf{w}}^{(epoch-1)}$ and randomly shuffle $n$ training instances in $\mathcal{D}$

\For{$t = 1,\dotsc,n$}{
   {\bfseries Sequentially pick:} $\{\mathbf{x}_{it},y_{it}\}$ from $\mathcal{D}$ , $it \in \{1,...,n\}$

   {\bfseries Compute:} $g(\mathbf{w}^{(t-1)}) = (\langle \mathbf{w}^{(t-1)},\mathbf{x}_{it}\rangle + b)y_{it}$

   {\bfseries Update:} $\mathbf{w}^{(t)} =\left\{
\begin{aligned}
I&: w^{(t-1)} - \eta\big[\lambda  \mathbf{w}^{(t-1)} - (1-s^*)\alpha_{s^*}\mathbf{x}_{it} y_{it}\frac{f(\alpha_{s^*},\beta_{s^*},g(\mathbf{w}^{(t-1)}))}{(1+f(\alpha_{s^*},\beta_{s^*},g(\mathbf{w}^{(t-1)})))^2}\big] \\
II&: \mathbf{w}^{(t-1)} - \eta\big[\lambda  \mathbf{w}^{(t-1)} - c^* \mathbf{x}_{it} y_{it}e^{f(c^*,g(\mathbf{w}^{(t-1)}))}\big]
\end{aligned}
\right.
$}
 {\bfseries option A:} $\tilde{\mathbf{w}}^{(epoch)} = \mathbf{w}^{(n)}$ or
 {\bfseries option B:} $\tilde{\mathbf{w}}^{(epoch)} = \frac{1}{n}\sum_{t=1}^{n}\mathbf{w}^{(t)}$
}
\KwOut{$\tilde{\mathbf{w}}^{(T_{max})}$}
\caption{Stochastic Gradient Descent with Robust Losses (\textbf{SGDRL})\label{SGDRL}}
\end{algorithm*}

\subsection{Convergence Analysis}\label{ConvergenceAnalysis}
When we apply SGD methods to SVM model with proposed robust losses, it converges into the qualified local minima. According to the \mbox{detailed} explanation about the three conditions in Section \ref{Losses and Algorithm}, the objective $G(\mathbf{w})$ satisfies the ARSC and $g_i(\mathbf{w})$ satisfies the ARSM. Based on the ARSC and ARSM, we can analyze the convergence rate of SGD methods using robust losses. We use $\mathbb{E}\big[\cdot\big]$ to denote the \mbox{expectation}.

\begin{theorem}\label{Convergence}
Consider that $G(\mathbf{w})$ satisfies Augmented Restricted Strong Convexity and $g_i(\mathbf{w})$ satisfies Augmented Restricted Smoothness. Define $\mathbf{w^*}$ as a local minima and $\beta$ as the parameter of Augmented Restricted Smoothness. Assume that learning rate $\eta$ is sufficient to let $G(\mathbf{w}^{(t)})$ be a non-increasing update. After $T$ iterations, we have
\begin{equation*}
G(\mathbf{w}^{(T)}) - G(\mathbf{w^*}) \leq \frac{\mathbb{E}\big[\lVert \mathbf{w}^{(0)} - \mathbf{w^*} \rVert^2\big]}{(2\eta - 12\eta^2\beta)\cdot T}
\end{equation*}
\end{theorem}
\subsection{Robustness Analysis}\label{RobustnessAnalysis}
Now we theoretically analyze the robustness of two representative robust losses. Assume that $\{\mathbf{x}_i,y_i\}_{i=1}^k$ is a random subset of the training data $\mathcal{D}$ and $f_\mathbf{w}$ is the decision function, according to the representer theorem, $z_i = y_if_\mathbf{w}(\mathbf{x}_i) = y_i(\sum_{j=1}^{k}\mathcal{K}(\mathbf{x}_j,\mathbf{x}_i)\alpha_j + b) = y_i \mathcal{K}_i^T\alpha + y_i b$, where $\alpha= (\alpha_1,\alpha_2,...,\alpha_k)'$, $\mathcal{K}= (\mathcal{K}_1,\mathcal{K}_2,...,\mathcal{K}_k)'$ and $\mathcal{K}_i=(\mathcal{K}(\mathbf{x}_1, \mathbf{x}_i),\mathcal{K}(\mathbf{x}_2, \mathbf{x}_i),...,\mathcal{K}(\mathbf{x}_k, \mathbf{x}_i))'$. $\lambda > 0$ is a regularizer parameter, $\mathcal{K}$ is a mercer kernel and $\mathcal{H_K}$ is a Reproducing Kernel Hilbert Space (RKHS). For a family of robust losses $r(z)$, we define two functions $\rho(z)$ and $\varrho(z)$ such that $r(z) = \rho(1-z)$ and $\varrho(z) = \frac{\rho'(z)}{z}$. According to the inference in supplementary materials, we define the weighted parameter $\phi_i$ as an important parameter that affects the update of the dual variable in SGD methods, where $\phi_i = \varrho(1 - y_i \mathcal{K}_i^T \alpha - y_i b)$. Moreover, $\phi_i$ is related to $\mathbf{x}_i$ for L2-SVM. We define $\delta = e^{\alpha_{s^*} \beta_{s^*}}$ for Smooth Ramp Loss. Therefore, the results of robustness analysis are provided in Theorem \ref{Robustness}. Due to the limit of space, the detailed proof of Theorem~\ref{Robustness} is provided in the supplementary materials.

\begin{theorem}\label{Robustness}
Assume that an instance $\mathbf{x}_i$ is annotated with noisy label $y_i$, which means $y_i (\mathcal{K}_i^T \alpha + b) <~0$. Its corresponding weighted parameter $\phi_i$ for Smooth Ramp Loss with $(s^*,\alpha_{s^*},\beta_{s^*})$ is
\begin{equation*}
\phi_i = \frac{(1-s^*)\alpha_{s^*}\delta e^{\alpha_{s^*}(y_i \mathcal{K}_i^T \alpha + y_i b)}}{(1 - (y_i \mathcal{K}_i^T \alpha + y_i b))(1 + \delta e^{\alpha_{s^*}(y_i \mathcal{K}_i^T \alpha + y_i b)})^2}
\end{equation*}
for Reversed Gompertz Loss with $c^*$ is
\begin{equation*}
\phi_i = \frac{c^*e^{c^*(y_i \mathcal{K}_i^T \alpha + y_i b)-e^{c^*(y_i \mathcal{K}_i^T \alpha + y_i b)}}}{1 - (y_i \mathcal{K}_i^T \alpha + y_i b)}
\end{equation*}
if $\vert f_w(\mathbf{x}_i) \vert = \vert (\mathcal{K}_i^T \alpha + b) \vert$ increases, which means $\mathbf{x}_i$ with noisy label $y_i$ becomes an outlier, then both $\phi_i$ will definitely decrease. It indicates that the proposed Robust Losses do reduce the negative effects introduced by noisy labels.
\end{theorem}
\section{Experiments}
In this section, we mainly perform experiments on noisy datasets to verify the convergence and robustness of SGD methods with two representative robust losses. The datasets \cite{chang2010training} range from small to large scale. For convenience, we \mbox{abbreviate} SGD with Smooth Ramp Loss as SGD(SRamp) and SGD with Reversed Gompertz Loss as SGD(RGomp) respectively.
\subsection{Experimental Settings}\label{Experimental Testbed and Setup}
All experimental datasets come from the LIBSVM datasets webpage \cite{CC01a}\footnote{ \url{http://www.csie.ntu.edu.tw/~cjlin/libsvmtools/datasets/}} . The statistics of the datasets are summarized in Table \ref{Datasets}. Among them, REAL-SIM, COVTYPE, MNIST38 and IJCNN1 are manually split into the training set and testing set by about $4:1$. We normalize the data by scaling each feature to $[0,1]$. To generate the datasets with noisy labels, we follow the settings in \cite{yang2012multiple,natarajan2013learning}. Specifically, we proportionally flip the class label of training data. For example, we randomly flip 20\% of data labels from \mbox{$-1$ to $1$ or $1$ to $-1$}, and assume that the data has $20\%$ of noisy labels. We then repeat the same process to produce 40\% and 60\% of noisy labels on all datasets.

In the experiments, the baseline methods are classified into two categories. The first category consists of SGD methods with different losses ranging from convex losses to robust nonconvex losses, which can verify the convergence and robustness of SGD methods with two representative losses for noisy labels. For example, we choose SGD with Logistic Loss (SGD(Log)), Hinge Loss (SGD(Hinge)) and Ramp Loss (SGD(Ramp)). We also choose ASGD \cite{xu2011towards} with Logistic Loss (ASGD(Log)) and PEGASOS \cite{ShalevShwartzSSC11} as baseline methods. For the second category, we compare proposed methods with LIBLINEAR \cite{REF08a} (We abbreviate L2-regularized L2-loss SVM Primal solution as LIBPrimal and Dual solution as LIBDual) due to its wide popularity in large-scale machine learning. All the methods are implemented in C++. Experiments are performed on a computer with a 3.20GHz Inter CPU and 8GB main memory running on a Windows 7.

The regularization parameter $\lambda$ is chosen by $10$-fold cross validation for all methods in the range of \{$10^{-6},10^{-5},10^{-4},10^{-3},10^{-2},10^{-1},1,10$\}. For SGD methods with different losses, the number of epochs is normally set to $15$ for convergence comparison and the primal variable $\mathbf{w}$ is initialized to $\mathbf{0}$. For LIBLINEAR, we set the bias $b$ to $1$ and the stopping tolerance $\epsilon$ to $10^{-2}$ for primal solution and $10^{-1}$ for dual solution by default. For PEGASOS, the number of epochs for convergence is set to $\frac{10}{\lambda}$ by default and the block size $k$ is set to $1$ for training efficiency. For SGD(SRamp), the parameter $s^*$ is chosen by $10$-fold cross validation in the range of $[-2,0]$ according to real-world datasets. Therefore, the parameter $(s^*,\alpha_{s^*},\beta_{s^*})$ is optimized to $(-0.7,3,-0.15)$, $(-1,2,-0.03)$ or $(-2,1.5,0.5)$. For SGD(RGomp), the parameter $c^*$ is randomly fixed to~$2$.  All the experiments are repeated ten times and the results are averaged over the $10$ trials. Methods are indicated by \lq\lq-\rq\rq in Table~\ref{TER-Table} due to running out of memory. Methods are not reported in Figures \ref{Testing error rate} and \ref{Variance} due to running out of memory or too long training time.\footnote{On MNIST38 and SUSY datasets, PEGASOS run out of memory, and the training time of LIBDual is several orders of magnitude more than that of other baselines.}

\begin{table}[!tp]
\centering
\caption{Datasets used in the experiments.}\label{Datasets}
\begin{tabular}{|c|c|c|c|}
\hline
DATA SET & TRAINING PTS. & TESTING PTS. & FEATURES\\\hline
A7A & 16,100 & 16,461 & 123 \\\hline
IJCNN1 & 49,990 & 91,701 & 22\\\hline
REAL-SIM & 57,847 & 14,462 & 20,958\\\hline
COVTYPE & 464,810 & 116,202 & 54\\\hline
MNIST38 & 450,000 & 97,570 & 784\\\hline
SUSY & 4,000,000 & 1,000,000 & 18 \\
\hline
\end{tabular}
\end{table}

\subsection{The Performance of Convergence}\label{Performance on Noisy Data}
\begin{figure*}[!tp]
\begin{center}
\centerline{\includegraphics[width=1.2\textwidth]{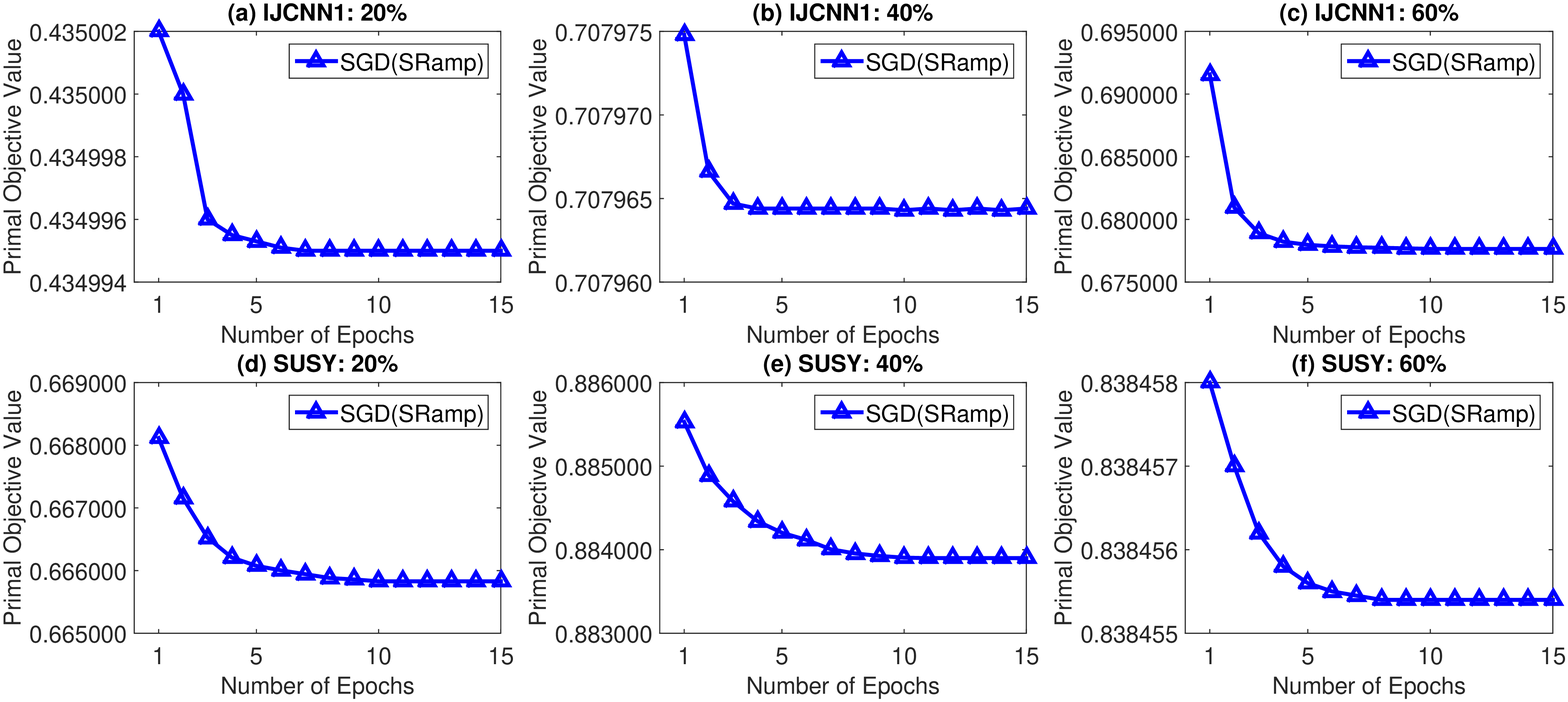}}
\caption{Primal objective value of SGD(SRamp) with the number of epochs on representative small-scale IJCNN1 and large-scale SUSY. Datasets have varying percentages (in \%) of noisy labels (20\%, 40\% and 60\%).}
\label{pov:demo}
\end{center}
%
\begin{center}
\centerline{\includegraphics[width=1.2\textwidth]{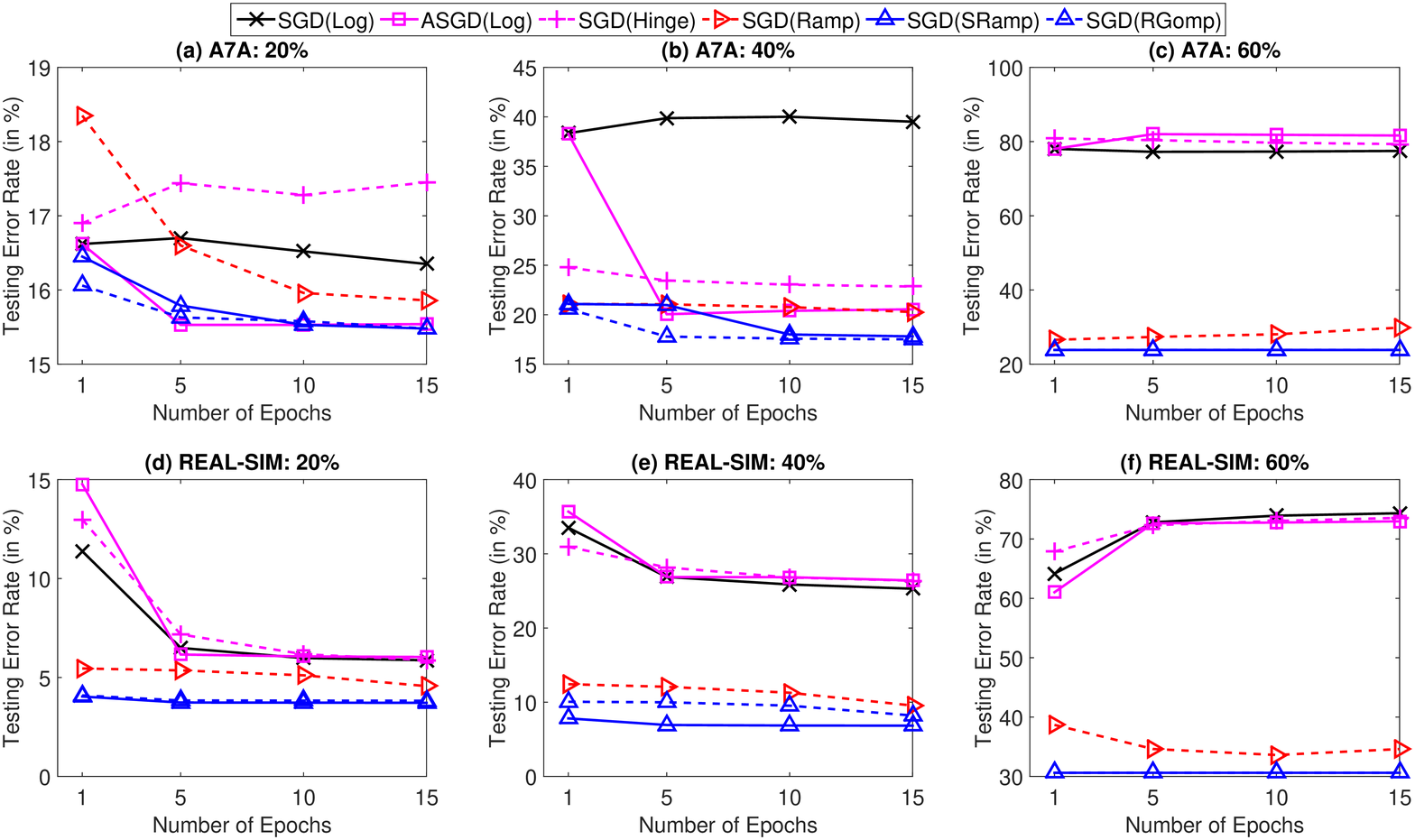}}
\caption{Testing error rate (in \%) with the number of epochs on A7A and REAL-SIM. Datasets have varying percentages (in \%) of noisy labels (20\%, 40\% and 60\%). For PEGASOS, the number of epochs for convergence is set to $\frac{10}{\lambda}$ by default. Therefore, we do not report its result.}
\label{TER-ITER}
\end{center}
\end{figure*}

First, we verify the convergence of SGD methods with two representative losses for noisy labels. Due to the limit of space, we only provide the primal objective value of SGD(SRamp) with the number of epochs on representative small-scale IJCNN1 and large-scale SUSY datasets. The two datasets have varying percentages of noisy labels. \mbox{Figure \ref{pov:demo}} shows the primal objective value of SGD(SRamp) with the number of epochs. We observe that SGD(SRamp) converges within $15$ epochs. This observation is consistent with our convergence analysis in Section \ref{ConvergenceAnalysis}. Since SGD(SRamp) and SGD(RGomp) are very similar, the convergence curve of SGD(RGomp) is also similar to that of SGD(SRamp). Thus, we do not report the results of SGD(RGomp).

Then, we further observe the convergence comparison of SGD methods with different losses for noisy labels in Figure \ref{TER-ITER} where, with the increase of number of epochs, the testing error rate of SGD(SRamp) and SGD(RGomp) not only decrease faster than that of other baseline methods but also keep relative stable in the most cases. By contrast, the convergence performance of SGD(Ramp) is not as good as our methods due to its nonsmoothness and nonconvexity.

\subsection{The Performance of Robustness}\label{Performance on Clean Data}
\begin{figure*}[!tp]
\begin{center}
\centerline{\includegraphics[width=1.2\textwidth]{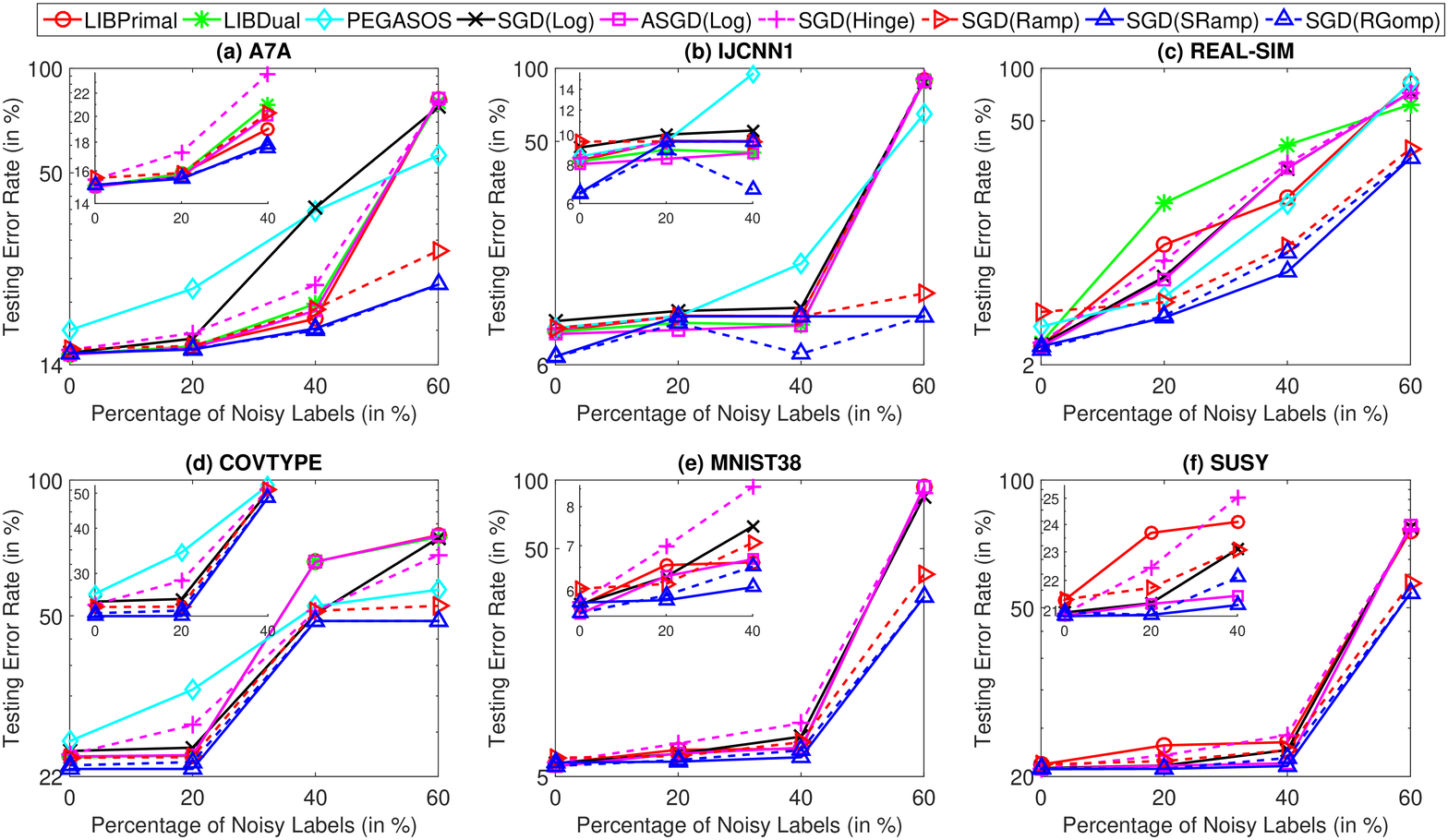}}
\caption{Testing error rate (in \%) on datasets with varying percentages (in \%) of noisy labels. We provide the subfigures to compare the testing error rate with 0\% to 40\% of noisy labels on all datasets except for REAL-SIM. The y-axis is in log-scale.}
\label{Testing error rate}
\end{center}
%
\begin{center}
\centerline{\includegraphics[width=1.2\textwidth]{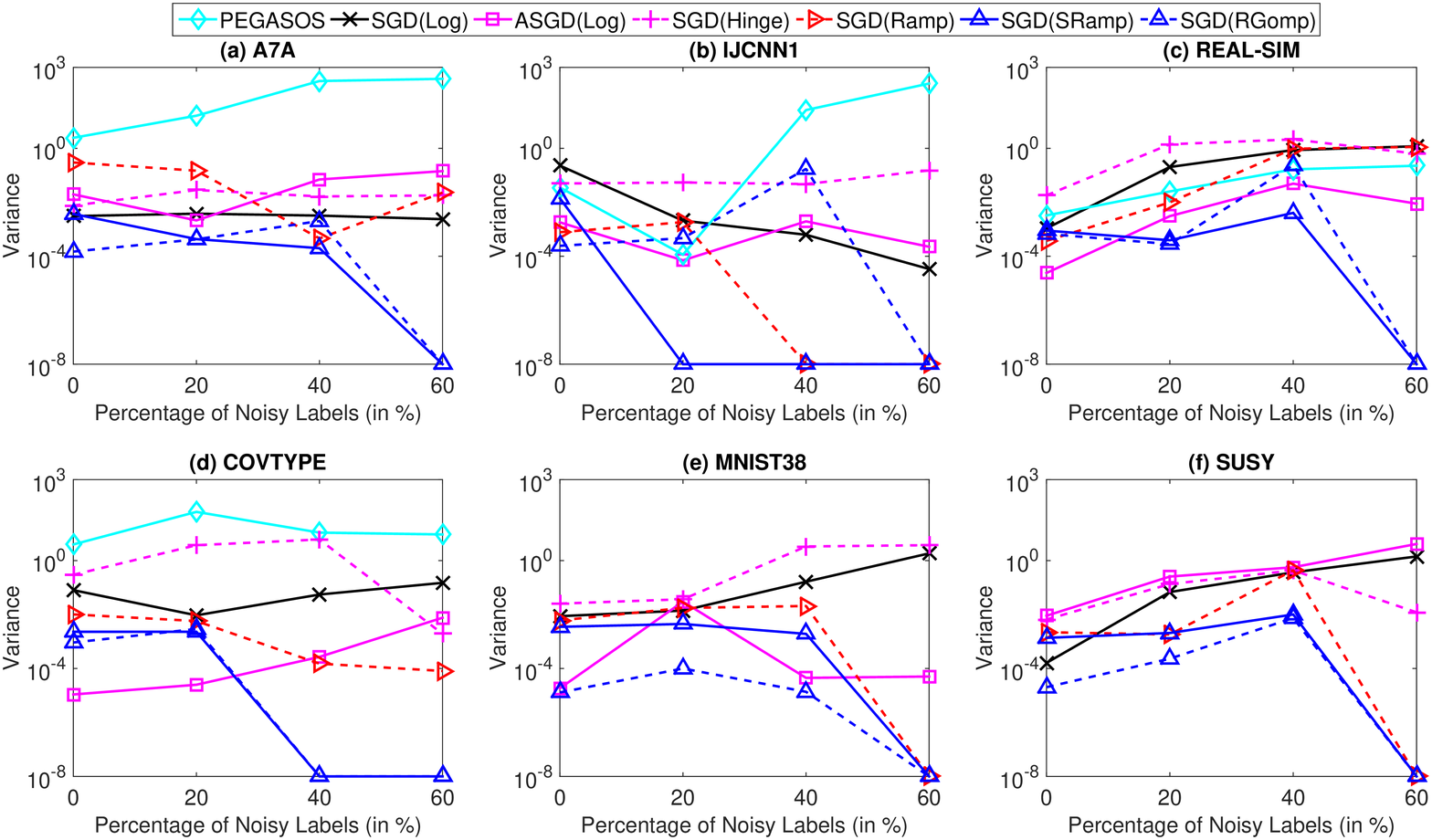}}
\caption{Variance on datasets with varying percentages (in \%) of noisy labels. The y-axis is in log-scale. Note that there is no variance for LIBPrimal and LIBDual because in each update of the primal variable, they compute full gradients instead of stochastic gradients.}
\label{Variance}
\end{center}
\vskip -0.3in
\end{figure*}

\begin{table}[!tp]
\centering
\caption{Testing error rate (in \%) with standard deviation on datasets without noisy labels. Methods are indicated by \lq\lq-\rq\rq due to running out of memory.}\label{TER-Table}
\begin{tabular}{|c|c|c|c|c|c|c|}
\hline
METHODS & A7A & IJCNN1 & REAL-SIM & COVTYPE & MNIST38 & SUSY\\\hline
LIBPRIMAL & 14.99 & 8.25 & 2.57 & 24.35 & 5.71 & 21.34\\\hline
LIBDUAL & \textbf{15.02} & 8.20 & 2.67 & 24.25 & 6.09 & 35.32\\\hline
PEGASOS & 17.62$\pm$1.56 & 8.50$\pm$0.19 & 3.32$\pm$0.06 & 26.36$\pm$1.99 & - & -\\\hline
SGD(Log) & 15.16$\pm$0.06 & 9.08$\pm$0.48 & 2.62$\pm$0.03 & 25.07$\pm$0.28 & 5.73$\pm$0.09 & 20.93$\pm$0.01\\\hline
ASGD(Log) & 14.99$\pm$0.14 & 8.04$\pm$0.04 & 2.54$\pm$0.01 & 24.38$\pm$0.01 & \textbf{5.54$\pm$0.01} & 20.83$\pm$0.09\\\hline
SGD(Hinge) & 15.45$\pm$0.09 & 8.40$\pm$0.22 & 2.69$\pm$0.13 & 24.62$\pm$0.54 & 5.77$\pm$0.16 & 20.89$\pm$0.08\\\hline
SGD(Ramp) & 15.54$\pm$0.54 & 8.50$\pm$0.03 & 4.02$\pm$0.02 & 24.22$\pm$0.10 & 6.04$\pm$0.08 & 21.36$\pm$0.05\\\hline
SGD(SRamp) & 15.11$\pm$0.06 & 6.49$\pm$0.12 & 2.55$\pm$0.03 & 23.69$\pm$0.04 & 5.76$\pm$0.06 & \textbf{20.81$\pm$0.03}\\\hline
SGD(RGomp) & 15.10$\pm$0.01 & \textbf{6.45$\pm$0.02} & \textbf{2.45$\pm$0.03} & \textbf{23.29$\pm$0.03} & 5.56$\pm$0.01 & 20.94$\pm$0.01\\
\hline
\end{tabular}
\end{table}

Finally, we verify the robustness of SGD methods with two representative losses for noisy labels. Figures \ref{Testing error rate} and \ref{Variance} respectively report testing error rate and variance with varying percentages of noisy labels. From Figures \ref{Testing error rate} and \ref{Variance}, we have the following observations. (a) On all datasets, SGD(SRamp) and SGD(RGomp) obviously outperform the other baseline methods in testing error rate beyond $40\%$ of noisy labels. Between $0\%$ to $40\%$, SGD(SRamp) and SGD(RGomp) still have comparative advantages. In particular, for a high-dimensional dataset REAL-SIM, the advantage of SGD(SRamp) and SGD(RGomp) is extremely obvious in the whole range of the x-axis. (b) Meanwhile, we notice that the variance of testing error rate for baseline methods (e.g., PEGASOS) gradually increases with the growing percentage of noisy labels, but the variance of testing error rate for SGD(SRamp) and SGD(RGomp) remains at the lowest level in the most cases. Therefore, the robustness of SGD(SRamp) and SGD(RGomp) have been validated by their testing error rate and variance.

In the most cases, the proposed SGD(SRamp) and SGD(RGomp) outperform other baseline methods not only on datasets with varying percentage of noisy labels but also on clean datasets. For example, Table \ref{TER-Table} demonstrates that in terms of the testing error rate with the standard deviation, SGD(SRamp) and SGD(RGomp) outperform other baseline methods on IJCNN1, REAL-SIM, COVTYPE and SUSY datasets without noisy labels.

\section{Conclusions}
This paper studies SGD methods with a family of robust losses for the label noise problem. For convenience, we mainly introduce two representative robust losses including Smooth Ramp Loss and Reversed Gompertz Loss. Our theoretical analysis not only reveals that the convergence rate is $\mathcal{O}(1/T)$ for SGD methods using robust losses, but also proves the robustness of two representative robust losses. Comprehensive experimental results show that, on real-world datasets with varying percentages of noisy labels, SGD methods
using our proposed losses are robust enough to reduce negative effects caused by noisy labels with fast convergence. In the future, we will extend our proposed robust losses to improve the performance of SGD methods for regression problems with noisy labels.


\end{document}